\documentclass[10pt,twocolumn,letterpaper]{article}

\usepackage{cvpr}              
\usepackage[accsupp]{axessibility}  
\usepackage{makecell}
\usepackage{multirow}
\usepackage{graphicx}

\usepackage[table]{xcolor}
\usepackage{xcolor}
\usepackage{appendix}
\usepackage{bm}

\definecolor{iccvblue}{rgb}{0.21,0.49,0.74}
\definecolor{cell}{RGB}{211,211,211}
\usepackage[pagebackref,breaklinks,colorlinks,allcolors=iccvblue]{hyperref}

\def\paperID{749} 
\def\confName{ICCV}
\def\confYear{2025}

\title{Rectifying Magnitude Neglect in Linear Attention}

\author{%
  Qihang Fan$^{1, 2}$, Huaibo Huang$^{1}$\thanks{Huaibo Huang is the corresponding author.}, Yuang Ai$^{1, 2}$, Ran He$^{1, 2}$\\
  $^1$MAIS \& NLPR, Institute of Automation, Chinese Academy of Sciences, Beijing, China\\
  $^2$School of Artificial Intelligence, University of Chinese Academy of Sciences, Beijing, China\\
  \texttt{fanqihang.159@gmail.com, huaibo.huang@cripac.ia.ac.cn, }\\
  \texttt{shallowdream555@gmail.com, rhe@nlpr.ia.ac.cn}
}

\begin{document}
\maketitle
\begin{abstract}
As the core operator of Transformers, Softmax Attention exhibits excellent global modeling capabilities. However, its quadratic complexity limits its applicability to vision tasks. In contrast, Linear Attention shares a similar formulation with Softmax Attention while achieving linear complexity, enabling efficient global information modeling. Nevertheless, Linear Attention suffers from a significant performance degradation compared to standard Softmax Attention. In this paper, we analyze the underlying causes of this issue based on the formulation of Linear Attention. We find that, unlike Softmax Attention, Linear Attention entirely disregards the magnitude information of the Query($Q$ or $\phi(Q)$). This prevents the attention score distribution from dynamically adapting as the Query scales. As a result, despite its structural similarity to Softmax Attention, Linear Attention exhibits a significantly different attention score distribution. Based on this observation, we propose \textbf{Magnitude-Aware Linear Attention} (MALA), which modifies the computation of Linear Attention to fully incorporate the Query’s magnitude. This adjustment allows MALA to generate an attention score distribution that closely resembles Softmax Attention while exhibiting a more well-balanced structure.  We evaluate the effectiveness of MALA on multiple tasks, including \textbf{image classification, object detection, instance segmentation, semantic segmentation, natural language processing, speech recognition, and image generation}. Our MALA achieves strong results on all of these tasks. Code will be available at \url{https://github.com/qhfan/MALA}.
\end{abstract}    
\section{Introduction}
\label{sec:intro}
\begin{figure}
    \centering
    \includegraphics[width=0.95\linewidth]{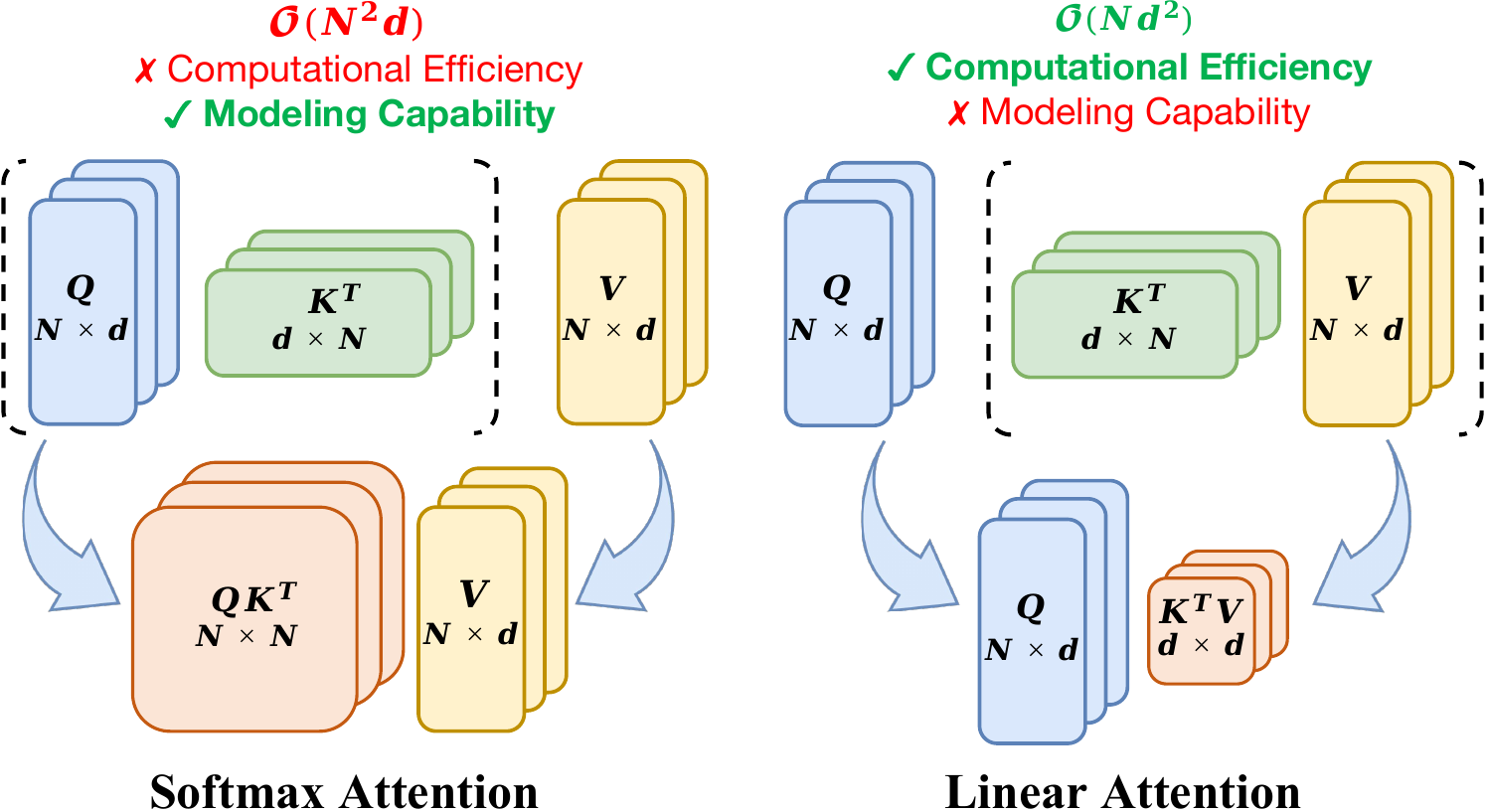}
    \vspace{-2mm}
    \caption{Comparison between Softmax Attention and Linear Attention. While linear attention offers linear complexity and high computational efficiency, its modeling capability falls short compared to Softmax Attention.}
    \vspace{-4mm}
    \label{fig:intro}
\end{figure}

Since the introduction of the Transformer~\cite{attention, vit} into the vision domain, it has gained increasing attention. Its exceptional global modeling capability has enabled Vision Transformers to achieve outstanding performance in various visual tasks, such as image classification, object detection, and semantic segmentation, fully demonstrating the Transformer’s potential in vision applications~\cite{flattentrans, SwinTransformer}.

\begin{figure*}
    \centering
    \includegraphics[width=0.95\linewidth]{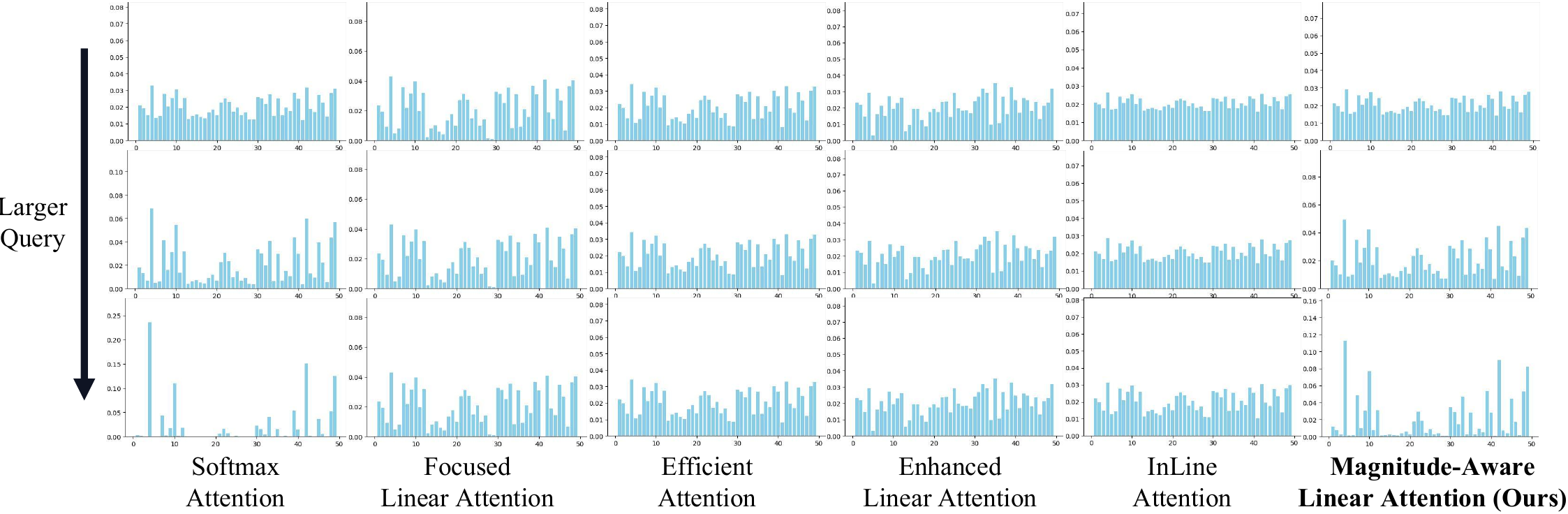}
    \vspace{-2mm}
    \caption{Comparison of attention score distributions across different mechanisms. As the magnitude of the Query ($Q$ or $\phi(Q)$) increases, the attention score distribution in Softmax Attention becomes increasingly spiky, concentrating more attention on keys that originally have higher scores. In contrast, Linear Attention maintains an unchanged distribution or exhibits only minimal variation, resulting in a relatively smooth attention score distribution. Our MALA retains the spiky characteristic of Softmax Attention while preventing it from becoming excessively sharp, achieving a more balanced distribution.}
    \vspace{-4mm}
    \label{fig:distribution}
\end{figure*}

However, the core operator of the Transformer, Softmax Attention, has a quadratic complexity with respect to the number of tokens $ N $, resulting in high computational costs that significantly hinder its widespread adoption in the vision domain. Many models reduce the computational cost of Softmax Attention by decreasing the number of tokens involved in its computation, bringing its complexity closer to or even achieving linearity~\cite{pvt, pvtv2, SwinTransformer, cswin, fan2023rmt, cmt}. However, these methods, which limit the number of tokens, often fail to accurately model the relationships between all tokens globally, preventing the Transformer from fully leveraging its original advantages.

Unlike these improvements to Softmax Attention, linear attention fundamentally eliminates the Softmax operation. As shown in Fig.~\ref{fig:intro}, by removing the Softmax operation, the computation order of $ Q $, \( K \), and \( V \) is rearranged, resulting in a linear complexity with respect to the number of tokens \( N \). Although Linear Attention and Softmax Attention share a very similar form, the removal of the Softmax operation introduces several challenges, often leading to significantly inferior performance compared to Softmax Attention.

In this paper, we analyze the computational formulation of Linear Attention and observe that it entirely disregards the magnitude information of the Query ($Q$ or $\phi(Q)$), preserving only its directional component. Consequently, Linear Attention exhibits a substantial discrepancy in attention score distribution compared to Softmax Attention. Specifically, as illustrated in Fig.~\ref{fig:distribution}, for a fixed direction, the attention scores in Softmax Attention become increasingly spiky as the Query magnitude increases, concentrating more attention on keys that originally have higher attention scores. In contrast, due to the inherent limitations of its computation, Linear Attention either maintains a fixed attention score distribution or undergoes only minimal variation, and the distribution remains consistently smooth. This phenomenon may account for its weak local perception and the tendency to produce overly smooth attention scores~\cite{efficientvit, flattentrans, devil, inline}.

To address this issue and better align the attention score distribution of Linear Attention with that of Softmax Attention, we propose Magnitude-Aware Linear Attention (MALA). MALA fully integrates the magnitude information of the Query, mimicking the variation trend of Softmax Attention while achieving a more balanced and reasonable allocation of attention. As a result, MALA outperforms Softmax Attention while preserving its linear complexity. To demonstrate the effectiveness of MALA, we conduct extensive experiments on image classification, object detection, instance segmentation, semantic segmentation, natural language processing, speech recognition, and image generation. Strong results across all these tasks demonstrate the effectiveness of proposed MALA. 


Our contributions can be summarized as follows:
\begin{itemize}
    \item We analyze the computational formulation of Linear Attention and reveal that it entirely disregards variations in the Query ($Q$ or $\phi(Q)$)’s magnitude. This omission leads to a substantial discrepancy between the attention score distributions of Linear Attention and Softmax Attention.  
    \item To bridge this gap, we propose Magnitude-Aware Linear Attention (MALA), which fully incorporates the Query’s magnitude information. MALA mimics the variation trend of Softmax Attention while achieving a more balanced and principled attention score distribution.  
    \item Based on MALA, we develop the Magnitude-Aware Vision Transformer (MAViT). We also test MALA on other tasks, such as natural language processing, speech recognition, and image generation. All models achieve promising results.
\end{itemize}

\section{Related Works}
\label{sec:related_works}
\paragraph{Vision Transformers.}Vision Transformer (ViT) is a powerful foundational vision model inspired by advancements in natural language processing (NLP)~\cite{vit, attention}. It demonstrates remarkable performance across various vision tasks. However, the core operator of the Transformer, Softmax Attention, has a quadratic complexity with respect to the number of tokens \( N \), imposing a computational burden that limits the application of Transformers in vision tasks.  Many works have proposed improvements to address this issue. One approach adopts a grouping strategy, where tokens are divided into multiple groups, reducing the computational burden at the cost of sacrificing the global receptive field of ViT~\cite{SwinTransformer, cswin, biformer, NAT, davit}. Another approach directly downsamples the tokens, preserving ViT's global perception capability but compromising its fine-grained representation~\cite{pvt, pvtv2, cmt, FAT, DGT, iformer}. Some methods integrate convolution with Transformers to enhance model's efficiency~\cite{uniformer, cloformer, structvit}. However, most of these approaches still rely on the quadratic complexity of Softmax Attention.  

\vspace{-3mm}

\paragraph{Linear Attention.}Linear Attention assumes that the exponential function can be approximated by the product of kernel functions. This decomposition reformulates the computation of attention scores, reducing the complexity of Attention to linear time. However, this improvement in efficiency comes at the cost of performance degradation. Many works have explored ways to bridge the gap between Softmax Attention and Linear Attention~\cite{flattentrans, SOFT, MLLA, efficientvit, efficientattn}. Among them, MILA~\cite{MLLA}, inspired by Mamba, incorporates Mamba's macro architecture into the design of Linear Attention. EfficientViT~\cite{efficientvit} and Flatten Transformer~\cite{flattentrans} integrate Linear Attention with convolution to compensate for its limitations in capturing local features. In contrast to these methods, we directly address the computational form of Linear Attention and the distribution of attention scores, aiming to align the behavior of Linear Attention with that of Softmax Attention.

\section{Method}
\label{sec:method}
\subsection{Preliminary}
Given an input token sequence \( X \in \mathbb{R}^{N \times d} \) of length \( N \) and dimension \( d \), the output of the \( i \)th token \( X_i \) after attention processing can be expressed as:
\begin{equation}
\begin{aligned}
    &Q=XW_Q, K=XW_K, V=XW_V, \\
    &Y_i=\sum_{j=1}^N\frac{{\rm Sim}(Q_i, K_j)}{\sum_{m=1}^N {\rm Sim}(Q_i, K_m)}V_{j};
\end{aligned}
\end{equation}
Where $W_Q,W_K,W_V \in \mathbb{R}^{d\times d}$ are learnable matrices, ${\rm Sim}(.,.)$ is the similarity function. In classical Softmax Attention, ${\rm Sim}(Q_i,K_j)={\rm exp}(Q_iK^T_j/\sqrt{d})$. This requires computing the exponential value for each pair of query and key, resulting in a complexity of $O(N^2)$. 

In Linear Attention, this situation changes. It employs a kernel function \( \phi(.) \) to approximate the similarity function and maps \( Q \) and \( K \) into positive real numbers. and leading to ${\rm Sim}(Q_i,K_j)=\phi(Q_i)\phi(K_j)^T$. Based on this transformation, the formulation of Linear Attention can be rewritten as:
\begin{equation}
    \begin{aligned}
    Y_i &= \sum_{j=1}^N\frac{\phi(Q_i)\phi(K_j)^T}{\sum_{m=1}^N \phi(Q_i)\phi(K_m)^T}V_j \\
    & = \frac{\phi(Q_i)(\sum_{j=1}^{N}\phi(K_j)^TV_j)}{\phi(Q_i)(\sum_{m=1}^N \phi(K_m)^T)};
    \end{aligned}
\end{equation}
in this computational form, the order of operations for $Q$, $K$, and $V$ changes from $(QK^T)V$ to $Q(K^TV)$, eliminating the need to compute the result for each query-key pair. This reduces the complexity with respect to the number of tokens $N$ from $O(N^2)$ to $O(N)$. However, the reduction in complexity also leads to a decline in performance.  

\subsection{Magnitude Neglect in Linear Attention}
We define  
\begin{equation}
    \phi(Q_i) = \|\phi(Q_i)\| \vec{\bm{\alpha}_i};
\end{equation}
where \(\|\phi(Q_i)\|\) represents the magnitude of \(\phi(Q_i)\), and \(\vec{\bm{\alpha_i}}\) denotes its direction vector. Substituting this expression into the formulation of Linear Attention, we obtain:
\begin{equation}
    \begin{aligned}
    Y_i & = \frac{||\phi(Q_i)||\vec{\bm{\alpha}_i}(\sum_{j=1}^{N}\phi(K_j)^TV_j)}{||\phi(Q_i)||\vec{\bm{\alpha}_i}(\sum_{m=1}^N\phi(K_m)^T)} \\
    & = \frac{\vec{\bm{\alpha}_i}(\sum_{j=1}^{N}\phi(K_j)^TV_j)}{\vec{\bm{\alpha}_i}(\sum_{m=1}^N\phi(K_m)^T)};
    \end{aligned}
\end{equation}
from this equation, we observe that the magnitude information of \(\phi(Q)\) in Linear Attention is completely ignored. As a result, as long as \(\vec{\bm{\alpha}}\) remains fixed, the attention score distribution of Linear Attention remains unchanged. 

This phenomenon leads to a significant discrepancy between the attention score distributions of Linear Attention and Softmax Attention. In Softmax Attention, the magnitude of \( Q_i \) is fully taken into account. Given \( Q_i \), the ratio of its attention scores for two different keys, \( K_m \) and \( K_n \), is given by
\begin{equation}
    \label{eq:ratio}
    \frac{{\rm exp}(Q_i K_m^T/\sqrt{d})}{{\rm exp}(Q_i K_n^T/\sqrt{d})}=p;
\end{equation}
We assume that \( Q_i \) assigns a higher attention weight to \( K_m \), i.e., \( p > 1 \). When the direction of \( Q_i \) remains unchanged and its magnitude is scaled by a factor of \( a > 1 \), the ratio of its attention scores for \( K_m \) and \( K_n \) becomes: 
\begin{equation}
    \label{eq:ratiop}
    \frac{{\rm exp}(aQ_iK_m^T/\sqrt{d})}{{\rm exp}(aQ_iK_n^T/\sqrt{d})}=\frac{{\rm exp}(Q_i K_m^T/\sqrt{d})^a}{{\rm exp}(Q_i K_n^T/\sqrt{d})^a}=p^a=p_{s};
\end{equation}
Since \( p > 1 \) and \( a > 1 \), it follows that \( p_s > p \). Given that the attention scores of \( Q_i \) across all \( K \)s sum to 1, Eq.~\ref{eq:ratio} and Eq.~\ref{eq:ratiop} imply that \textbf{as the magnitude \( \|Q_i\| \) increases, the attention of \( Q_i \) becomes more concentrated on keys with higher original attention scores, while the attention assigned to keys with lower initial scores diminishes.}

However, this situation does not occur in Linear Attention. The ratio of \( Q_i \)'s attention to \( K_m \) and \( K_n \) is given by:
\begin{equation}
    \frac{\phi(Q_i)\phi(K_m)^T}{\phi(Q_i)\phi(K_n)^T}=\frac{||\phi(Q_i)||\vec{\bm{\alpha}_i}\phi(K_m)^T}{||\phi(Q_i)||\vec{\bm{\alpha}_i}\phi(K_n)^T}=\frac{\vec{\bm{\alpha}_i}\phi(K_m)^T}{\vec{\bm{\alpha}_i}\phi(K_n)^T};
\end{equation}
This indicates that \textbf{regardless of the changes in the magnitude of \( \phi(Q_i) \), the attention scores in Linear Attention remain in the same distribution and do not concentrate on specific keys.} This distinction explains why the attention scores learned by Linear Attention are less spiky compared to those of Softmax Attention and why the learned features exhibit weaker locality~\cite{devil, flattentrans, inline, efficientvit}. 

\begin{figure}
    \centering
    \includegraphics[width=0.95\linewidth]{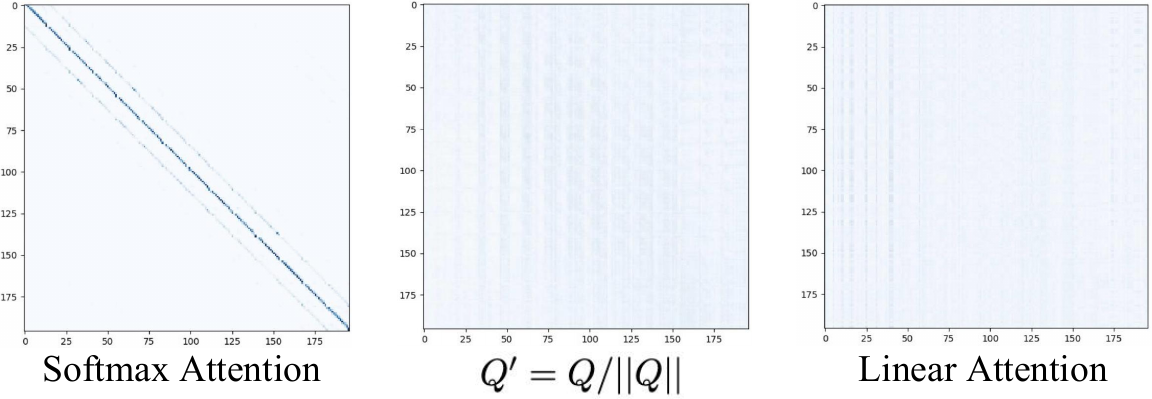}
    \vspace{-2mm}
    \caption{Attention scores of different models. When $Q$ is replaced with $Q/||Q||$, Softmax Attention exhibits a distribution similar to that of Linear Attention, becoming much smoother and losing locality.}
    \vspace{-4mm}
    \label{fig:pre}
\end{figure}

In addition to the theoretical analysis above, we also conduct experimental validation. As shown in Tab.~\ref{tab:absoft}, based on DeiT-T, we rewrite the $Q$ in Softmax Attention as $Q/||Q||$, thereby disregarding the magnitude information. We observe a significant drop in the model's performance, which becomes similar to that of the model based on Linear Attention. We visualize the attention scores in Fig.~\ref{fig:pre} and find that the distribution converges to that of Linear Attention, becoming much smoother and losing locality.

\begin{table}[t]
    \centering
    \scalebox{0.8}{
    \begin{tabular}{c|c| c c}
    \toprule[1pt]
         Model & Softmax & $Q'=Q/||Q||$ & Softmax$\xrightarrow{}$Linear \\
         Acc(\%) & 72.2 & 70.0 & 69.8 \\
    \bottomrule[1pt]
    \end{tabular}}
    \vspace{-2mm}
    \caption{Discarding magnitude information in Softmax Attention.}
    \vspace{-4mm}
    \label{tab:absoft}
\end{table}

\subsection{Magnitude-Aware Linear Attention}
To bridge the gap between Linear Attention and Softmax Attention, we aim for Linear Attention to incorporate the magnitude information \( \|\phi(Q_i)\| \) and exhibit similar variation trend as Softmax Attention.  

\begin{figure}[t]
    \centering
    \includegraphics[width=0.85\linewidth]{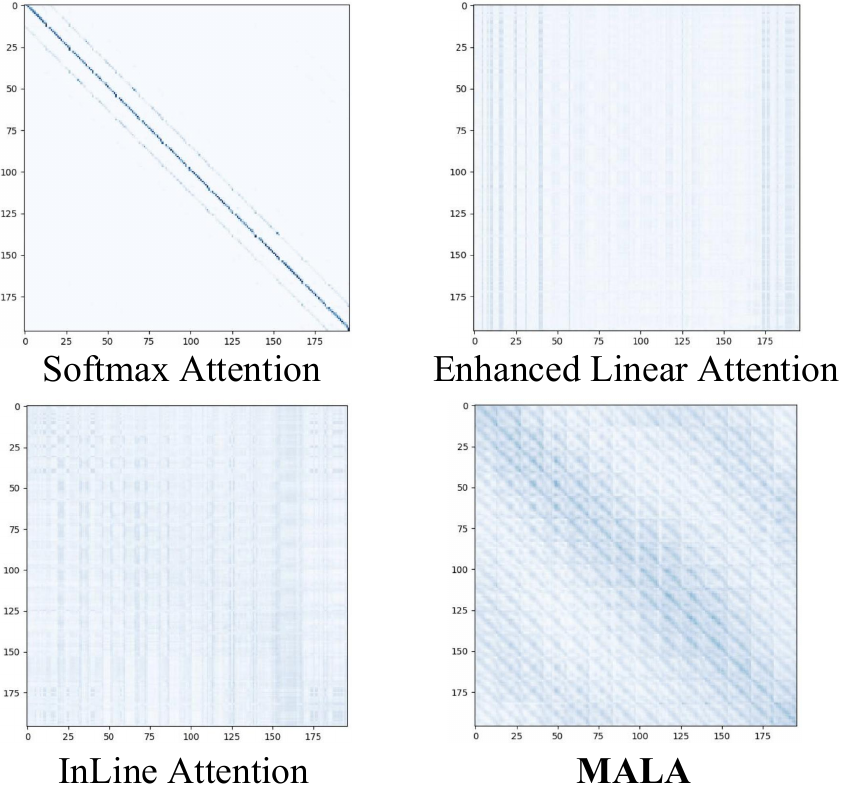}
    \vspace{-2mm}
    \caption{Visualization of attention scores on DeiT-T setting. Softmax Attention's score is too spiky and primarily focuses on local regions, while Linear Attention's score is too smooth and excessively disregards local information. In contrast, MALA effectively balances both aspects. The \textcolor{red}{visualizations on natural images} are provided in the \textcolor{red}{Appendix}.}
    \vspace{-4mm}
    \label{fig:attnmap}
\end{figure}


In our Magnitude-Aware Linear Attention(MALA), building upon the original Linear Attention, we introduce a scaling factor and an offset term while discarding the division-based normalization in favor of an addition-based normalization:  
\begin{equation}
    {\rm Attn}(Q_i, K_j)=\beta\phi(Q_i)\phi(K_j)^T-\gamma;
\end{equation}
Where:
\begin{equation}
\begin{aligned}
    &\beta=1+\frac{1}{\phi(Q_i)\sum_{m=1}^N\phi(K_m)^T},\\
    &\gamma=\frac{\phi(Q_i)\sum_{m=1}^N\phi(K_m)^T}{N},\\
    \sum_{j=1}^{N}{\rm Attn}&(Q_i,K_j)=\beta\sum_{j=1}^N\phi(Q_i)\phi(K_j)^T-N\gamma=1;\\
\end{aligned}
\end{equation}
When considering all attention scores as positive values, the ratio of \( Q_i \)'s attention scores for \( K_m \) and \( K_n \) is given by: 
\begin{equation}
    \frac{\beta\phi(Q_i)\phi(K_m)^T-\gamma}{\beta\phi(Q_i)\phi(K_n)^T-\gamma}=p;
\end{equation}
We assume that \( Q_i \) assigns a higher attention score to \( K_m \), i.e., \( \beta\phi(Q_i)\phi(K_m)>\beta\phi(Q_i)\phi(K_n) \) and $p>1$.When the direction of \( \phi(Q_i) \) remains unchanged and its magnitude is scaled by a factor of \( a > 1 \), it is straightforward to derive that the new \( \beta \) and \( \gamma \) can be written as: 
\begin{equation}
\begin{aligned}
    \beta_{new}&=\frac{\beta+a-1}{a}, \\
    \gamma_{new}&=a\gamma;
\end{aligned}
\end{equation}
At this point, the ratio of the attention scores of \( a\phi(Q_i) \) for \( K_m \) and \( K_n \) becomes: 
\begin{equation}
\label{eq:malaratio}
\begin{aligned}
    &\frac{\beta_{new}a\phi(Q_i)\phi(K_m)^T-\gamma_{new}}{\beta_{new}a\phi(Q_i)\phi(K_n)^T-\gamma_{new}}\\
    =&\frac{\beta\phi(Q_i)\phi(K_m)^T-\frac{a\beta}{\beta+a-1}\gamma}{\beta\phi(Q_i)\phi(K_n)^T-\frac{a\beta}{\beta+a-1}\gamma}=p_{m};
\end{aligned}
\end{equation}
Since \( \beta > 1 \) and \( a > 1 \), it is straightforward to prove that \( \frac{a\beta}{a+\beta-1} > 1 \). From this, we can further easily prove that when considering all attention scores as positive values, \( p_{m} > p \) (Details can be found in the \textcolor{red}{Appendix}). Moreover, since \( \sum_{j=1}^{N} {\rm Attn}(Q_i, K_j) = 1 \), as the magnitude of \( \phi(Q_i) \) increases, MALA concentrates more attention on the keys that originally received higher attention, while allocating less attention to the keys that originally had lower attention. \textbf{This behavior is similar to Softmax Attention.}


Although both Softmax Attention and MALA exhibit a trend of more concentrated attention score distributions as the magnitude of \( Q_i \) or \( \phi(Q_i) \) increases, the rate at which this concentration occurs differs between the two. From the comparison between Eq.~\ref{eq:ratiop} and Eq.~\ref{eq:malaratio}, it can be observed that in Softmax Attention, the ratio \( p \) of attention scores exhibits an \textbf{exponential growth} with respect to the scaling factor \( a \) of \( \|Q\| \). In contrast, in MALA, the ratio \( p \) follows a \textbf{fractional growth} pattern with respect to the scaling factor \( a \) of \( \|\phi(Q)\| \). The variation of \( p \) in MALA is smaller than that in Softmax Attention, which may contribute to the superior performance of MALA over Softmax Attention. As shown in Fig.~\ref{fig:attnmap}, we visualize the attention scores of different mechanisms. It can be observed that Softmax Attention’s score is too spiky and primarily focuses on local regions. In contrast, Linear Attention’s score is too smooth and excessively disregards local information~\cite{inline, flattentrans, efficientvit}. MALA, however, effectively balances both aspects. This indicates that the gradual variation of \( p \) in MALA leads to a more appropriately distributed attention score. 

As for the occurrence of negative/zero attention scores in MALA, in our experiments (image classification, object detection, instance segmentation and semantic segmentation), we find that although negative/zero attention scores are theoretically possible, their actual frequency of occurrence is equal zero. We do not observe any negative or zero attention scores. So we do not introduce additional considerations.

When the attention scores are applied to the values, the complete formulation of MALA is expressed as: 
\begin{equation}
    \begin{aligned}
        Y_i&=\sum_{j=1}^N(\beta\phi(Q_i)\phi(K_j)^T-\gamma)Vj\\
        &=\beta\phi(Q_i)\sum_{j=1}^N\phi(K_j)^TV_j-\gamma\sum_{j=1}^NVj;
    \end{aligned}
\end{equation}
Where:
\begin{equation}
\begin{aligned}
    &\beta=1+\frac{1}{\phi(Q_i)\sum_{m=1}^N\phi(K_m)^T},\\
    &\gamma=\frac{\phi(Q_i)\sum_{m=1}^N\phi(K_m)^T}{N};\\
\end{aligned}
\end{equation}


\section{Experiments}
\label{sec:exp}
\begin{table}[ht]
    \centering
    \setlength{\tabcolsep}{1.6mm}
    \subfloat{
    \scalebox{0.76}{
    \begin{tabular}{c|c|c|c c|c}
    \toprule[1pt]
         Cost & Model & Type & \makecell{Parmas\\(M)} & \makecell{FLOPs\\(G)} & \makecell{Top1-acc\\(\%)}\\
         \midrule[0.5pt]
         \multirow{8}{*}{\rotatebox{90}{\makecell{Tiny model\\$\sim 2.5$G}}} 
         & NAT-M~\cite{NAT} & Trans & 20 & 2.7 & 81.8 \\ 
         & FAT-B2~\cite{FAT} & Trans & 14 & 2.0  & 81.9 \\
         & GC-ViT-XT~\cite{globalvit} & Trans & 20 & 2.6 & 82.0 \\
         & RMT-T~\cite{fan2023rmt} & Trans & 14 & 2.5 & 82.4 \\
         & MSVMamba-M~\cite{msvmamba} & Mamba & 12 & 1.5 & 79.8 \\
         & Flatten-PVTv2-B1~\cite{flattentrans} & Linear & 13 & 2.2 & 79.5 \\
         & RAVLT-T~\cite{RALA} & Linear & 15 & 2.4 & 82.8 \\
         & \cellcolor{cell}MAViT-T & \cellcolor{cell}Linear & \cellcolor{cell}16 & \cellcolor{cell}2.5 & \cellcolor{cell}\textbf{82.9} \\
         \midrule[0.5pt]
         \multirow{11}{*}{\rotatebox{90}{\makecell{Small model\\$\sim 4.5$G}}}
         & MogaNet-S~\cite{iclr2024MogaNet} & CNN & 25 & 5.0 & 83.4 \\
         & SG-Former-S~\cite{sgformer} & Trans & 23 & 4.8 & 83.2 \\
         & FAT-B3~\cite{FAT} & Trans & 29 & 4.4 & 83.6 \\
         & SMT-S~\cite{SMT} & Trans & 20 & 4.8 & 83.7 \\
         & RMT-S~\cite{fan2023rmt} & Trans & 27 & 4.5 & 84.1 \\
         & SECViT-S~\cite{fan2024semantic} & Trans & 27 & 4.6 & 84.3 \\
         & Vmamba-T~\cite{vmamba} & Mamba & 30 & 4.9 & 82.6 \\
         & MSVMamba-T~\cite{msvmamba} & Mamba & 32 & 5.1 & 83.0 \\
         & Flatten-CSwin-T~\cite{flattentrans} & Linear & 21 & 4.3 & 83.1 \\
         & RAVLT-S~\cite{RALA} & Linear & 26 & 4.6 & 84.4 \\
         & \cellcolor{cell}MAViT-S & \cellcolor{cell}Linear & \cellcolor{cell}27 & \cellcolor{cell}4.6 & \cellcolor{cell}\textbf{84.7} \\
         \midrule[0.5pt]
         \multirow{11}{*}{\rotatebox{90}{\makecell{Base model\\$\sim 10.0$G}}}
         & ConvNeXT-S~\cite{convnext} & CNN & 50 & 8.7 & 83.1 \\
         & InternImage-S~\cite{internimage} & CNN & 50 & 8.0 & 84.2 \\
         & MogaNet-B~\cite{iclr2024MogaNet} & CNN & 44 & 9.9 & 84.3 \\
         & BiFormer-B~\cite{biformer} & Trans & 57 & 9.8 & 84.3 \\
         & RMT-B~\cite{fan2023rmt} & Trans & 54 & 9.7 & 85.0 \\
         & SECViT-B~\cite{fan2024semantic} & Trans & 57 & 9.8 & 85.2 \\
         & Vmamba-S~\cite{vmamba} & Mamba & 50 & 8.7 & 83.6 \\
         & MSVMamba-S~\cite{msvmamba} & Mamba & 50 & 8.8 & 84.1 \\
         & MILA-S~\cite{MLLA} & Linear & 43 & 7.3 & 84.4 \\
         & RAVLT-B~\cite{RALA}& Linear & 48 & 9.9 & 85.5 \\
         & \cellcolor{cell}MAViT-B & \cellcolor{cell}Linear & \cellcolor{cell}50 & \cellcolor{cell}9.9 & \cellcolor{cell}\textbf{85.7} \\
         \midrule[0.5pt]
         \multirow{11}{*}{\rotatebox{90}{\makecell{Large model\\$\sim 15.0$G}}}
         & MogaNet-L~\cite{iclr2024MogaNet} & CNN & 83 & 15.9 & 84.7 \\
         & InterImage-B~\cite{internimage} & CNN & 97 & 16.0 & 84.9 \\
         & SG-Former-B~\cite{sgformer} & Trans & 78 & 15.6 & 84.7 \\
         & STViT-L~\cite{stvit} & Trans & 95 & 15.6 & 85.3 \\
         & RMT-L~\cite{fan2023rmt} & Trans & 95 & 18.2 & 85.5 \\
         & Vmamba-B~\cite{vmamba} & Mamba & 89 & 15.4 & 83.9 \\
         & MSVMamba-B~\cite{msvmamba} & Mamba & 91 & 16.3 & 84.4 \\
         & SOFT-Huge~\cite{SOFT} & Linear & 87 & 16.3 & 83.3 \\
         & InLine-CSwin-B~\cite{inline} & Linear & 73 & 14.9 & 84.5 \\
         & RAVLT-L~\cite{RALA} & Linear & 95 & 16.0 & 85.8 \\
         & \cellcolor{cell}MAViT-L & \cellcolor{cell}Linear & \cellcolor{cell}98 & \cellcolor{cell}16.1 & \cellcolor{cell}\textbf{86.0} \\
         \bottomrule[1pt]
    \end{tabular}
    }}
    \vspace{-2mm}
    \caption{Comparison with the state-of-the-art on ImageNet-1K classification. We use "CNN" to refer to convolutional neural networks, "Trans" to refer to Vision Transformers, "Mamba" to refer to visual state space model, and "Linear" to refer to models based on Linear Attention. }
    \vspace{-4mm}
    \label{tab:IN1K}
\end{table}

\begin{table}[t]
    \centering
    \setlength{\tabcolsep}{0.22mm}
    \scalebox{0.75}{
    \begin{tabular}{c|c|c c|c c c c c c}
        \toprule[1pt]
         Backbone & Type & \makecell{Params\\(M)} & \makecell{FLOPs\\(G)} & $AP^b$ & $AP^b_{50}$ & $AP^b_{75}$ & $AP^m$ & $AP^m_{50}$ & $AP^m_{75}$\\
          \midrule[0.5pt]
          \multicolumn{10}{c}{Mask R-CNN $3\times$+MS} \\
          \midrule[0.5pt]
          NAT-T~\cite{NAT} & Trans & 48 & 258 & 47.8 & 69.0 & 52.6 & 42.6 & 66.0 & 45.9 \\
          SMT-S~\cite{SMT} & Trans & 40 & 265 & 49.0 & 70.1 & 53.4 & 43.4 & 67.3 & 46.7\\
          RMT-S~\cite{fan2023rmt} & Trans & 46 & 262 & 50.7 & 71.9 & 55.6 & 44.9 & 69.1 & 48.4\\
          Vmamba-T~\cite{vmamba} & Mamba & 50 & 271 & 48.8 & -- & -- & 43.7 & -- & -- \\
          MILA-T~\cite{MLLA} & Linear & 44 & 255 & 48.8 & 71.0 & 53.6 & 43.8 & 68.0 & 46.8 \\
          \rowcolor{cell}MAViT-S & Linear & 44 & 262 & \textbf{51.4} & \textbf{72.6} & \textbf{56.2} & \textbf{45.5} & \textbf{69.8} & \textbf{49.2}\\
          \midrule[0.5pt]
          InternImage-S~\cite{internimage} & CNN & 69 & 340 & 49.7 & 71.1 & 54.5 & 44.5 & 68.5 & 47.8 \\
          SMT-B~\cite{SMT} & Trans & 52 & 328 & 49.8 & 71.0 & 54.4 & 44.0 & 68.0 & 47.3\\
          RMT-B~\cite{fan2023rmt} & Trans & 73 & 373 & 52.2 & 72.9 & 57.0 & 46.1 & 70.4 & 49.9  \\
          Vmamba-S~\cite{vmamba} & Mamba & 70 & 349 & 49.9 & -- & -- & 44.2 & -- & -- \\
          MILA-S~\cite{MLLA} & Linear & 63 & 319 & 50.5 & 71.8 & 55.2 & 44.9 & 69.1 & 48.2 \\
          \rowcolor{cell}MAViT-B & Linear & 67 & 372 & \textbf{53.2} & \textbf{74.1} & \textbf{58.5} & \textbf{47.0} & \textbf{71.5} & \textbf{51.1} \\
          \midrule[0.5pt]
          InternImage-B~\cite{internimage} & CNN & 115 & 501 & 50.3 & 71.4 & 55.3 & 44.8 & 68.7 & 48.0 \\
          Swin-B~\cite{SwinTransformer} & Trans & 107 & 496 & 48.6 & 70.0 & 53.4 & 43.3 & 67.1 & 46.7 \\
          CSwin-B~\cite{cswin} & Trans & 97 & 526 & 50.8 & 72.1 & 55.8 & 44.9 & 69.1 & 48.3 \\
          \rowcolor{cell}MAViT-L & Linear & 114 & 501 & \textbf{53.6} & \textbf{74.3} & \textbf{58.7} & \textbf{47.2} & \textbf{71.5} & \textbf{51.4} \\
          \midrule[0.5pt]
          \multicolumn{10}{c}{Cascade Mask R-CNN $3\times$+MS}\\
          \midrule[0.5pt]
          HorNet-T~\cite{hornet} & CNN & 80 & 728 & 52.4 & 71.6 & 56.8 & 45.6 & 69.1 & 49.6 \\
          GC-ViT-T~\cite{globalvit} & Trans & 85 & 770 & 51.6 & 70.4 & 56.1 & 44.6 & 67.8 & 48.3 \\
          CSWin-T~\cite{cswin} & Trans & 80 & 757 & 52.5 & 71.5 & 57.1 & 45.3 & 68.8 & 48.9 \\
          RMT-S~\cite{fan2023rmt} & Trans & 83 & 741 & 53.2 & 72.0 & 57.8 & 46.1 & 69.8 & 49.8\\
          FL-Swin-T~\cite{flattentrans} & Linear & 87 & 747 & 50.8 & 69.6 & 55.1 & 44.1 & 67.0 & 48.1 \\
          \rowcolor{cell}MAViT-S & Linear & 82 & 741 & \textbf{54.2} & \textbf{72.6} & \textbf{58.6} & \textbf{47.0} & \textbf{70.5} & \textbf{51.1} \\
          \midrule[0.5pt]
          NAT-S~\cite{NAT} & Trans & 108 & 809 & 51.9 & 70.4 & 56.2 & 44.9 & 68.2 & 48.6 \\
          UniFormer-B~\cite{uniformer} & Trans & 107 & 878 & 53.8 & 72.8 & 58.5 & 46.4 & 69.9 & 50.4 \\
          RMT-B~\cite{fan2023rmt} & Trans & 111 & 852 & 54.5 & 72.8 & 59.0 & 47.2 & 70.5 & 51.4  \\
          FL-Swin-S~\cite{flattentrans} & Linear & 108 & 841 & 52.2 & 71.2 & 56.8 & 45.4 & 68.3 & 49.4 \\
          InLine-Swin-S~\cite{inline} & Linear & 109 & 835 & 52.4 & 71.0 & 56.9 & 45.4 & 68.8 & 49.6 \\
          \rowcolor{cell}MAViT-B & Linear & 105 & 851 & \textbf{55.5} & \textbf{74.0} & \textbf{60.4} & \textbf{48.0} & \textbf{71.7} & \textbf{52.5}\\
          \midrule[0.5pt]
          ConvNeXt-B~\cite{convnext} & CNN & 145 & 964 & 52.7 & 71.3 & 57.2 & 45.6 & 68.9 & 49.5\\
          Swin-B~\cite{SwinTransformer} & Trans & 145 & 982 & 51.9 & 70.9 & 56.5 & 45.0 & 68.4 & 48.7\\
          CSwin-B~\cite{cswin} & Trans & 135 & 1004 & 53.9 & 72.6 & 58.5 & 46.4 & 70.0 & 50.4 \\
          \rowcolor{cell}MAViT-L & Linear & 152 & 979 & \textbf{56.0} & \textbf{74.6} & \textbf{60.9} & \textbf{48.4} & \textbf{72.4} & \textbf{52.9} \\
          \bottomrule[1pt]
    \end{tabular}}
    \vspace{-2mm}
    \caption{Comparison to other backbones on "$3\times+\mathrm{MS}$" schedule.}
    \vspace{-4mm}
    \label{tab:COCO3x}
\end{table}

\begin{table*}[t]
    \setlength{\tabcolsep}{1.25mm}
    \centering
    \scalebox{0.75}{
    \begin{tabular}{c|c|c c|c c c c c c|c c|c c c c c c}
        \toprule[1pt]
        \multirow{2}{*}{Backbone} & Type & \multirow{2}{*}{\makecell{Params\\(M)}} & \multirow{2}{*}{\makecell{FLOPs\\(G)}} & \multicolumn{6}{c|}{Mask R-CNN $1\times$} & \multirow{2}{*}{\makecell{Params\\(M)}} & \multirow{2}{*}{\makecell{FLOPs\\(G)}} & \multicolumn{6}{c}{RetinaNet $1\times$}\\
         & & & & $AP^b$ & $AP^b_{50}$ & $AP^b_{75}$ & $AP^m$ & $AP^m_{50}$ & $AP^m_{75}$ & & & $AP^b$ & $AP^b_{50}$ & $AP^b_{75}$ & $AP^b_S$ & $AP^b_{M}$ & $AP^b_{L}$ \\
         \midrule[1pt]
        PVTv2-B1~\cite{pvtv2} & Trans & 33 & 243 & 41.8 & 54.3 & 45.9 & 38.8 & 61.2 & 41.6 & 23 & 225 & 41.2 & 61.9 & 43.9 & 25.4 & 44.5 & 54.3 \\
        MPViT-XS~\cite{mpvit} & Trans & 30 & 231 & 44.2 & 66.7 & 48.4 & 40.4 & 63.4 & 43.4 & 20 & 211 & 43.8 & 65.0 & 47.1 & 28.1 & 47.6 & 56.5 \\
        FAT-B2~\cite{FAT} & 33 & 215 & 45.2 & 67.9 & 49.0 & 41.3 & 64.6 & 44.0 & 23 & 196 & 44.0 & 65.2 & 47.2 & 27.5 & 47.7 & 58.8 \\
        MSVmamba-M~\cite{msvmamba} & Mamba & 32 & 201 & 43.8 & 65.8 & 47.7 & 39.9 & 62.9 & 42.9 & -- & -- & -- & -- & -- & -- \\
        \rowcolor{cell}MAViT-T & Linear & 33 & 219 & \textbf{47.6} & \textbf{69.5} & \textbf{52.5} & \textbf{42.9} & \textbf{66.5} & \textbf{46.4} & 24 & 201 & \textbf{45.6} & \textbf{66.7} & \textbf{48.9} & \textbf{28.9} & \textbf{49.7} & \textbf{61.1} \\
        \midrule[1pt]
        CMT-S~\cite{cmt} & Trans & 45 & 249 & 44.6 & 66.8 & 48.9 & 40.7 & 63.9 & 43.4 & 44 & 231 & 44.3 & 65.5 & 47.5 & 27.1 & 48.3 & 59.1 \\
        FAT-B3~\cite{FAT} & 49 & -- & 47.6 & 69.7 & 52.3 & 43.1 & 66.4 & 46.2 & 39 & -- & 45.9 & 66.9 & 49.5 & 29.3 & 50.1 & 60.9 \\
        RMT-S~\cite{fan2023rmt} & Trans & 46 & 262 & 49.0 & 70.8 & 53.9 & 43.9 & 67.8 & 47.4 & 36 & 244 & 47.8 & 69.1 & 51.8 & 32.1 & 51.8 & 63.5 \\
        VMamba-T~\cite{vmamba} & Mamba & 50 & 271 & 47.3 & 69.3 & 52.0 & 42.7 & 66.4 & 45.9 & -- & -- & -- & -- & -- & -- \\
        MILA-T~\cite{MLLA} & Linear & 44 & 255 & 46.8 & 69.5 & 51.5 & 42.1 & 66.4 & 45.0 & -- & -- & -- & -- & -- & -- & -- & -- \\
        \rowcolor{cell}MAViT-S & Linear & 44 & 262 & \textbf{50.2} & \textbf{71.7} & \textbf{55.3} & \textbf{44.7} & \textbf{68.7} & \textbf{47.9} & 34 & 244 & \textbf{48.3} & \textbf{69.4} & \textbf{52.2} & \textbf{31.8} & \textbf{52.6} & \textbf{64.0} \\
        \midrule[1pt]
        CSWin-S~\cite{cswin} & Trans & 54 & 342 & 47.9 & 70.1 & 52.6 & 43.2 & 67.1 & 46.2 & -- & -- & -- & -- & -- & -- & -- & -- \\
        STViT-B~\cite{stvit} & Trans & 70 & 359 & 49.7 & 71.7 & 54.7 & 44.8 & 68.9 & 48.7 & -- & -- & -- & -- & -- & -- & -- & -- \\
        MSVmamba-S~\cite{mamba} & Mamba & 70 & 349 & 48.1 & 70.1 & 52.8 & 43.2 & 67.3 & 46.5 & -- & -- & -- & -- & -- & -- & -- & -- \\
        SOFT++-medium~\cite{Soft222} & Linear & 69 & 342 & 46.6 & 67.8 & 51.2 & 42.0 & 64.8 & 45.2 & 59 & 322 & 44.3 & 64.7 & 47.4 & 29.0 & 48.2 & 59.9 \\
        MLLA-S~\cite{MLLA} & Linear & 63 & 319 & 49.2 & 71.5 & 53.9 & 44.2 & 68.5 & 47.2 & -- & -- & -- & -- & -- & -- & -- & -- \\
        \rowcolor{cell}MAViT-B & Linear & 67 & 372 & \textbf{51.7} & \textbf{73.3} & \textbf{57.0} & \textbf{46.1} & \textbf{70.6} & \textbf{50.1} & 57 & 353 & \textbf{49.9} & \textbf{71.1} & \textbf{53.8} & \textbf{33.7} & \textbf{54.5} & \textbf{65.5}\\
        \midrule[1pt]
        InternImage-B~\cite{internimage} & CNN & 115 & 501 & 48.8 & 70.9 & 54.0 & 44.0 & 67.8 & 47.4 & -- & -- & -- & -- & -- & -- & -- & -- \\
        MPViT-B~\cite{mpvit} & Trans & 95 & 503 & 48.2 & 70.0 & 52.9 & 43.5 & 67.1 & 46.8 & 85 & 482 & 47.0 & 68.4 & 50.8 & 29.4 & 51.3 & 61.5 \\
        RMT-L~\cite{fan2023rmt} & Trans & 114 & 557 & 51.6 & 73.1 & 56.5 & 45.9 & 70.3 & 49.8 & 104 & 537 & 49.4 & 70.6 & 53.1 & 34.2 & 53.9 & 65.2 \\
        MILA-B~\cite{MLLA} & Linear & 115 & 502 & 50.5 & 72.0 & 55.4 & 45.0 & 69.3 & 48.6 & -- & -- & -- & -- & -- & -- & -- & -- \\
        \rowcolor{cell}MAViT-L & Linear & 114 & 501 & \textbf{52.5} & \textbf{73.6} & \textbf{57.8} & \textbf{46.5} & \textbf{71.0} & \textbf{50.6} & 104 & 482 & \textbf{50.6} & \textbf{71.7} & \textbf{54.9} & \textbf{34.1} & \textbf{55.3} & \textbf{65.6} \\
        \bottomrule[1pt]
    \end{tabular}}
    \vspace{-2mm}
    \caption{Comparison to other backbones with ``$1\times$" schedule.}
    \vspace{-4mm}
    \label{tab:COCO1x}
\end{table*}

We conduct extensive experiments on image classification, object detection, instance segmentation, semantic segmentation, natural language processing, speech recognition and image generation. Additionally, we perform ablation studies to validate the impact of MALA. More \textcolor{red}{visualization results} and details can be found in the \textcolor{red}{Appendix}.

\subsection{Image Classification}
\noindent\textbf{Settings. }We follow the same training strategy in previous works with the only supervision being classification loss~\cite{deit, stvit, fan2023rmt, biformer, MLLA, cswin}. We train our models on ImageNet-1K~\cite{imagenet} from scratch.  The maximum rates of increasing stochastic depth~\cite{droppath} are set to 0.1/0.15/0.4/0.55 for MAViT-T/S/B/L, respectively. The batch size is set to 1024 and the max learning rate is 1e-3. We train all models for 300 epochs.

\noindent\textbf{Results. }we compare the performance of various models in Tab.~\ref{tab:IN1K}. Under models of comparable size, MAViT achieves the best results. Specifically, with 98M parameters and 16.1G FLOPs, MAViT-L achieves the accuracy of 86.0\%. This performance surpasses MILA, another Linear Attention method, by 0.7\%. Moreover, MAViT-S achieves an accuracy of 84.7\% with only 27M parameters and 4.6G FLOPs, surpassing the larger MILA-S.  

\subsection{Object Detection and Instance Segmentation}
\noindent\textbf{Settings. }Following previous works~\cite{fan2023rmt, SwinTransformer, biformer}, we use RetinaNet~\cite{retinanet}, Mask-RCNN~\cite{maskrcnn} and Cascade Mask R-
CNN~\cite{cai18cascadercnn} to evaluate our models. we use the commonly used "$1\times$" (12 epochs) setting for the RetinaNet and Mask R-CNN and "$3\times$+MS" (36 epochs) for Mask R-CNN and Cascade Mask R-CNN. 

\noindent\textbf{Results. }We show the experimental results in Tab.~\ref{tab:COCO3x} ("$3\times$+MS") and Tab.~\ref{tab:COCO1x} ("$1\times$"). MAViT demonstrates significant advantages over other models based on Linear Attention. Moreover, it surpasses models utilizing Softmax Attention across all model scales. Specifically, MAViT-B achieves 55.5$AP^b$ and 48.0$AP^m$, which even surpass the larger CSwin-B (53.9$AP^b$ and 46.4$AP^m$) under the framework of Cascade Mask R-CNN.

\subsection{Semantic Segmentation}

\noindent\textbf{Settings. }Follow previous works~\cite{SwinTransformer, iformer, cmt}, we adopt SemanticFPN~\cite{semanticfpn} and UperNet~\cite{upernet} to evaluate our models. For SemanticFPN, we train the models for 80K iterations~\cite{pvt, pvtv2}, while for UperNet, we train them for 160K iterations. The batch sizes are set to 16 for all models. The input images are cropped to $512\times 512$. 

\noindent\textbf{Results. }We present the results in the Tab.~\ref{tab:seg}. MAViT surpasses other models across various sizes. Specifically, MAViT-B achieves 52.8 mIoU under the framework of UperNet, which surpasses larger MILA. MAViT-L can even achieve 53.6 mIoU. 

\begin{table}[ht]
    \centering
    \setlength{\tabcolsep}{0.45mm}
    \scalebox{0.75}{
    \begin{tabular}{c|c|c c c|c c c}
    \toprule[1pt]
    \multirow{3}{*}{Model} & \multirow{3}{*}{Type} & \multicolumn{3}{c|}{Semantic FPN 80K} & \multicolumn{3}{c}{Upernet 160K} \\
    & & \makecell{Params\\(M)} & \makecell{FLOPs\\(G)} & \makecell{mIoU\\(\%)} & \makecell{Params\\(M)} & \makecell{FLOPs\\(G)} & \makecell{mIoU$_{ss}$\\(\%)} \\
    \midrule[0.5pt]
    VAN-B1~\cite{VAN} & CNN & 18 & 140 & 42.9 & -- & -- & -- \\
    PVTv2-B1~\cite{pvtv2} & Trans & 18 & 136 & 42.5 & -- & -- & -- \\
    RMT-T~\cite{fan2023rmt} & Trans & 17 & 136 & 46.4 & -- & -- & -- \\
    MSVmamba-M~\cite{msvmamba} & Mamba & -- & -- & -- & 42 & 875 & 45.1 \\
    \rowcolor{cell}MAViT-T & Linear & 18 & 136 & \textbf{47.6} & 44 & 893 & \textbf{48.4} \\
    \midrule[0.5pt]
    MogaNet-S~\cite{iclr2024MogaNet} & CNN & 29 & 189 & 47.7 & 55 & 946 & 49.2 \\
    SMT-S~\cite{SMT} & Trans & -- & -- & -- & 50 & 935 & 49.2 \\
    RMT-S~\cite{fan2023rmt} & Trans & 30 & 180 & 49.4 & 56 & 937 & 49.8 \\
    Vmamba-T~\cite{vmamba} & Mamba & -- & -- & -- & 62 & 949 & 47.9 \\
    Fl-Swin-T~\cite{flattentrans} & Linear & -- & -- & -- & 60 & 946 & 44.8 \\
    \rowcolor{cell}MAViT-S & Linear & 28 & 180 & \textbf{50.7} & 55 & 937 & \textbf{51.0} \\
    \midrule[0.5pt]
    MogaNet-B~\cite{iclr2024MogaNet} & CNN & -- & -- & -- & 74 & 1050 & 50.1 \\
    RMT-B~\cite{fan2023rmt} & Trans & 57 & 294 & 50.4 & 83 & 1051 & 52.0 \\
    Vmamba-S~\cite{vmamba} & Mamba & -- & -- & -- & 82 & 1028 & 50.6 \\
    FL-Swin-S~\cite{flattentrans} & Linear & -- & -- & -- & 82 & 1038 & 48.1 \\
    \rowcolor{cell}MAViT-B & Linear & 51 & 292 & \textbf{51.5} & 77 & 1050 & \textbf{52.8} \\
    \midrule[0.5pt]
    MogaNet-L~\cite{iclr2024MogaNet} & CNN & -- & -- & -- & 113 &1176 & 50.9 \\
    CSWin-B~\cite{cswin} & Trans & 81 & 464 & 49.9 & 109 & 1222 & 51.1 \\
    SGFormer-B~\cite{sgformer} & Trans & 81 & 475 & 50.6 & 109 & 1304 & 52.0 \\
    RMT-L~\cite{fan2023rmt} & Trans & 98 & 482 & 51.4 & 125 & 1241 & 52.8 \\
    Vmamba-B~\cite{vmamba} & Mamba & -- & -- & -- & 122 & 1170 & 51.0 \\
    MILA-B~\cite{MLLA} & Linear & -- & -- & -- & 128 & 1183 & 51.9 \\
    \rowcolor{cell}MAViT-L & Linear & 98 & 424 & \textbf{52.8} & 125 & 1182 & \textbf{53.6} \\
    \bottomrule[1pt]

    \end{tabular}}
    \vspace{-2mm}
    \caption{ Comparison with the state-of-the-art on ADE20K.}
    \vspace{-6mm}
    \label{tab:seg}
\end{table}

\subsection{Inference Efficiency}
\noindent\textbf{Settings. }To thoroughly assess the inference efficiency of MAViT, we evaluate its performance on both low-resolution task (image classification with a resolution of \(224 \times 224\)) and high-resolution task (semantic segmentation with a resolution of \(512 \times 2048\)). For the low-resolution task, we measure model throughput using a batch size of 64. For the high-resolution task, we evaluate the frame per second (FPS) performance with a batch size of 1.

\begin{figure}[t]
    \centering
    \includegraphics[width=0.9\linewidth]{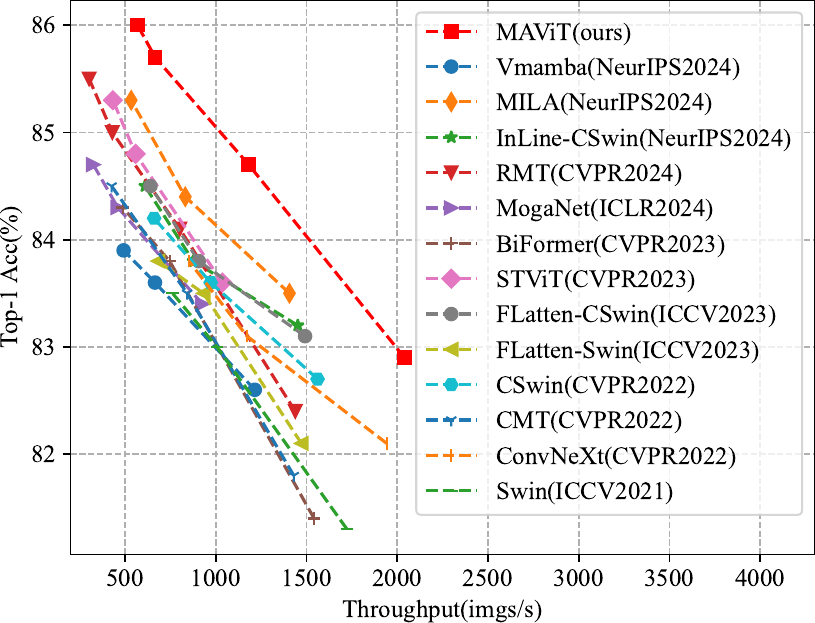}
    \vspace{-2mm}
    \caption{Comparison of general backbones’ inference speed on low resolution task (image classification, resolution $224\times 224$). The inference speed are measured on A100, batch size 64.}
    \vspace{-4mm}
    \label{fig:eff224}
\end{figure}
\begin{figure}[t]
    \centering
    \includegraphics[width=0.9\linewidth]{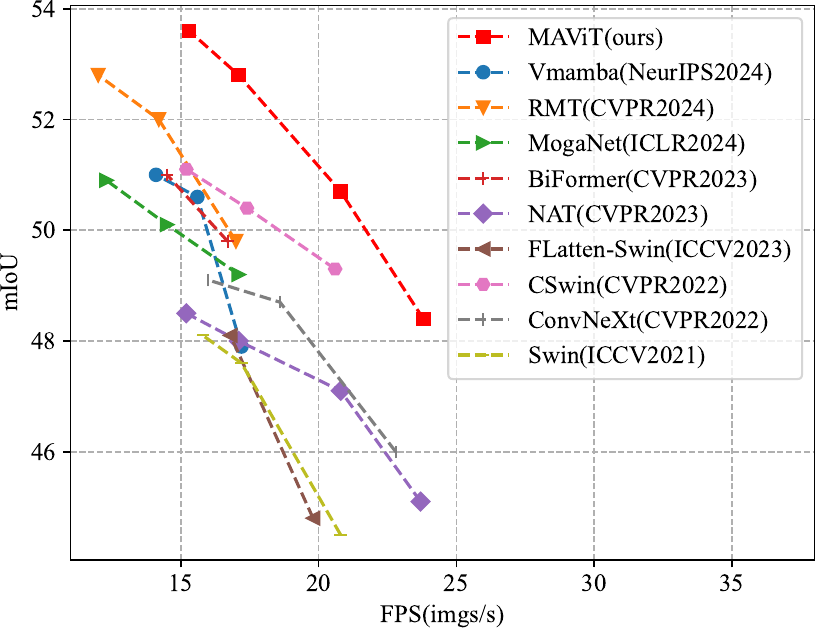}
    \vspace{-2mm}
    \caption{Comparison of general backbones’ inference speed on high resolution task (semantic segmentation with UperNet, resolution $512\times 2048$). The inference speed are measured on A100, batch size 1.}
    \vspace{-6mm}
    \label{fig:eff512}
\end{figure}

\noindent\textbf{Results. }The results are presented in Fig.~\ref{fig:eff224} and Fig.~\ref{fig:eff512}. Fig.~\ref{fig:eff224} illustrates the inference efficiency of different models on the low-resolution task, where MAViT achieves the best balance between throughput and accuracy. Similarly, for high-resolution tasks, the results in Fig.~\ref{fig:eff512} further highlight MAViT's superior efficiency. This demonstrates that MALA not only has a significantly lower theoretical complexity than Softmax Attention but also achieves high inference speed in practice.  

\subsection{Natural Language Processing}
\noindent\textbf{Settings.} Following previous works, we train the 0.3B MALA based model on 15B tokens, and evaluate the model on several commonly used benchmarks.

\noindent\textbf{Results. }We show the results in the Tab.~\ref{tab:nlp}. In the four commonly used benchmarks (LMB, PIQA, Hella, and Wino), our MALA exhibits strong performance.
\begin{table}[t]
    \centering
    \setlength{\tabcolsep}{1.0mm}
    \begin{tabular}{c|cccc}
    \toprule[1pt]
         & LMB$\uparrow$ & PIQA$\uparrow$ & Hella$\uparrow$ & Wino$\uparrow$ \\ 
         \midrule[0.5pt]
         Transformer~\cite{attention} & 31.0 & 63.3 & 34.0 & 50.4 \\
         RetNet~\cite{retnet} & 28.6 & 63.5 & 33.5 & 52.5 \\
         GLA~\cite{gla2024} & 30.3 & 64.8 & 34.5 & 51.4 \\
         Mamba~\cite{mamba} & 30.6 & 65.0 & 35.4 & 50.1 \\
         \midrule[0.5pt]
         \textbf{MALA} & \textbf{31.0} & \textbf{65.0} & \textbf{34.5} & \textbf{51.9} \\
         \bottomrule[1pt]
    \end{tabular}
    \vspace{-2mm}
    \caption{MALA in NLP.}
    \vspace{-4mm}
    \label{tab:nlp}
\end{table}

\subsection{Speech Recognition}
\noindent\textbf{Settings.} Our evaluation on speech recognition is based on the previous Conformer~\cite{conformer}. We replace the Softmax Attention in the Conformer with 1) vanilla Linear Attention and 2) our MALA. All training settings are the same as Conformer.

\begin{table}[ht]
    \centering
    \vspace{-3mm}
    \setlength{\tabcolsep}{0.6mm}
    \scalebox{0.88}{
    \begin{tabular}{cc|cccc}
    \toprule[1pt]
         \multirow{2}{*}{Model} & \multirow{2}{*}{Params} & \multicolumn{2}{c}{WER Without LM} & \multicolumn{2}{c}{WER With LM} \\
         & & testclean$\downarrow$ & testother$\downarrow$ & testclean$\downarrow$ & testother$\downarrow$ \\
         \midrule[0.5pt]
         Conformer(S) & 10.3 & 2.7 & 6.3 & 2.1 & 5.0 \\
         \midrule[0.5pt]
         Linear Attn & 10.3 & 3.4 & 10.2 & 2.6 & 7.3 \\
         MALA & 10.3 & \textbf{2.4} & \textbf{5.3} & \textbf{1.9} & \textbf{4.2} \\
         \bottomrule[1pt]
    \end{tabular}}
    \vspace{-3mm}
    \caption{MALA in speech recognition.}
    \vspace{-5mm}
    \label{tab:abspeech}
\end{table}

\noindent\textbf{Results.} We show the results in Tab.~\ref{tab:abspeech}. The results demonstrate that MALA perform better than Softmax Attention and vanilla Linear Attention.

\subsection{Image Generation}
\noindent\textbf{Settings.} Following previous works~\cite{DiT, ai2025dico, ai2025rela}, we train the models for 400K iterations with the batch size of 256 and learning rate of 1e-4.
\begin{table}[ht]
    \centering
    \setlength{\tabcolsep}{0.8mm}
    \scalebox{0.86}{
    \begin{tabular}{c|cc|cc}
    \toprule[1pt]
         Model & FLOPs & Throughput$\uparrow$ & FID$\downarrow$ & IS$\uparrow$ \\
         \midrule[0.5pt]
         DiT-S/2(400K)~\cite{DiT} & 250$\times$6.06G & 4.9imgs/s & 68.40 & -- \\
         DiG-S/2(400K)~\cite{DiG} & 250$\times$4.30G & 3.8imgs/s & 62.06 & 22.81 \\
         DiC-S/2(400K)~\cite{DiC} & 250$\times$5.90G & -- & 58.68 & 25.82 \\
         \midrule[0.5pt]
         MALA (400K) & \textbf{250$\times$4.26G} & \textbf{5.6imgs/s} & \textbf{49.62} & \textbf{32.18} \\
    \bottomrule[1pt]
    \end{tabular}}
    \vspace{-3mm}
    \caption{MALA for diffusion.}
    \vspace{-2mm}
    \label{tab:diffusion}
\end{table}

\noindent\textbf{Results.} We show the results in Tab.~\ref{tab:diffusion}. Compare to other methods based on ConvNet/Transformer, our model based on MALA exhibits better performance and faster speed, which demonstrate the superiority of MALA.

\subsection{Ablation Study}
\label{sec:ab}
\noindent\textbf{Comparison with Other Linear Attentions. }To ensure a fair comparison with previous state-of-the-art Linear Attention mechanisms, we adopt three model settings: DeiT-T, Swin-T, and Swin-S. Under these settings, we replace all instances of Softmax Attention with our proposed Magnitude-Aware Linear Attention while keeping all other components unchanged to maintain absolute fairness. The results are presented in Tab.~\ref{tab:strcomp}.  From the results, we observe that MALA achieves a significant improvement over previous linear attention mechanisms. Specifically, under the Swin-S setting, MALA outperforms InLine Attention by +1.7 in accuracy.  
\begin{table}[t]
    \centering
    \setlength{\tabcolsep}{1.1mm}
    \scalebox{0.75}{
    \begin{tabular}{c|c c|c}
    \toprule[1pt]         
         Linear Attention & Params(M) & FLOPs(G) & Top1-acc(\%)\\
         \midrule[0.5pt]
         \multicolumn{4}{c}{Comparison on DeiT-T Setting} \\
         \midrule[0.125pt]
         DeiT-T~\cite{deit} & 6 & 1.1 & 72.2 \\
         Hydra Attn~\cite{hydraattn} & 6 & 1.1 & 68.3 \\
         Efficient Attn~\cite{efficientattn} & 6 & 1.1 & 70.2 \\
         Linear Angular Attn~\cite{you2023castling} & 6 & 1.1 & 70.8 \\
         Enhanced Linear Attn~\cite{efficientvit} & 6 & 1.1 & 72.9 \\
         Focused Linear Attn~\cite{flattentrans} & 6 & 1.1 & 74.1 \\
         InLine Attn~\cite{inline} & 7 & 1.1 & 74.5 \\
         \rowcolor{cell} Magnitude-Aware Linear Attn & 6 & 1.1 & \textbf{75.1} \\
         \midrule[0.5pt]
         \multicolumn{4}{c}{Comparison on Swin-T Setting} \\
         \midrule[0.125pt]
         Swin-T~\cite{SwinTransformer} & 29 & 4.5 & 81.3 \\
         Hydra Attn~\cite{hydraattn} & 29 & 4.5 & 80.7 \\
         Efficient Attn~\cite{efficientattn} & 29 & 4.5 & 81.0 \\
         Linear Angular Attn~\cite{you2023castling} & 29 & 4.5 & 79.4 \\
         Enhanced Linear Attn~\cite{efficientvit} & 29 & 4.5 & 81.8 \\
         Focused Linear Attn~\cite{flattentrans} & 29 & 4.5 & 82.1 \\
         InLine Attn~\cite{inline} & 30 & 4.5 & 82.4 \\
         \rowcolor{cell}Magnitude-Aware Linear Attn & 29 & 4.5 & \textbf{83.7} \\
         \midrule[0.5pt]
         \multicolumn{4}{c}{Comparison on Swin-S Setting} \\
         \midrule[0.125pt]
         Swin-S~\cite{SwinTransformer} & 50 & 8.7 & 83.0 \\
         Focused Linear Attn~\cite{flattentrans} & 51 & 8.7 & 83.5 \\
         InLine Attn~\cite{inline} & 50 & 8.7 & 83.6 \\
         \rowcolor{cell}Magnitude-Aware Linear Attn & 50 & 8.7 & \textbf{85.3} \\
         \bottomrule[1pt]
    \end{tabular}}
    \vspace{-2mm}
    \caption{Comparison of different Linear Attentions based on DeiT-T, Swin-T, and Swin-S. MALA surpasses others by a large margin.}
    \vspace{-2mm}
    \label{tab:strcomp}
\end{table}

\begin{table}[t]
    \centering
    \scalebox{0.95}{
    \begin{tabular}{c|c c c}
    \toprule[1pt]
         $\phi(.)$ & ${\rm Elu}(.)+1$ & ${\rm ReLU}(.)$ & ${\rm exp}(.)$ \\
         \midrule[0.5pt]
         Acc(\%) & 82.9 & 82.8 & 82.9  \\
         mIoU & 47.6 & 47.7 & 47.4 \\
    \bottomrule[1pt]
    \end{tabular}}
    \vspace{-2mm}
    \caption{Effect of different kernel functions. }
    \vspace{-2mm}
    \label{tab:abkn}
\end{table}

\begin{table}[t]
    \centering
    \scalebox{0.95}{
    \begin{tabular}{c|c c c c}
    \toprule[1pt]
        Model & Acc(\%) & $AP^b$ & $AP^m$ & mIoU \\
        \midrule[0.5pt]
        MAViT-T & 82.9 & 47.6 & 42.9 & 47.6 \\
        \midrule[0.5pt]
        w/o $\beta$ & 52.3 & 24.6 & 18.7 & 22.2 \\
        w/o $\gamma$ & NaN & -- & -- & -- \\
        Learnable & 71.7 & 34.3 & 31.8 & 31.9 \\
    \bottomrule[1pt]
    \end{tabular}}
    \vspace{-2mm}
    \caption{Effect of $\beta$ and $\gamma$.}
    \vspace{-4mm}
    \label{tab:abbg}
\end{table}

\noindent\textbf{Kernel Function. }In MAViT, we employ $\phi(.) = \text{ELU}(.) + 1$ as the kernel function to ensure the non-negativity of $\phi(Q)$ and $\phi(K)$. To evaluate the impact of the kernel function, we conduct ablation studies based on MAViT-T. The results are presented in Tab.~\ref{tab:abkn}. Our MALA is not sensitive to the choice of kernel function, as almost any non-negative kernel function can achieve comparable performance.

\noindent\textbf{$\beta$ and $\gamma$.} $\beta$ and $\gamma$ are the core design elements of our model, endowing MALA with outstanding properties. We conduct ablation studies on these two parameters, and the results are presented in Tab.~\ref{tab:abbg}. We separately remove $\beta$ and $\gamma$, leading to a sharp decline in model performance, with some cases even resulting in NaN values. We also replace $\beta$ and $\gamma$ with learnable parameters, and the model's performance significantly deteriorates.

\section{Conclusion}
In this paper, we observe that the attention scores of Softmax Attention and Linear Attention exhibit distinct variation patterns as the magnitude of the Query ($Q$ or $\phi(Q)$) changes. From a formulation perspective, we analyze the underlying cause of this behavior and design Magnitude-Aware Linear Attention (MALA), which ensures that the attention scores of Linear Attention display a variation pattern similar to, yet more reasonable than, that of Softmax Attention. Based on MALA, we construct the Magnitude-Aware Vision Transformer (MAViT) and perform extensive experiments, which demonstrate the superior performance and high efficiency of MALA.

\section{Acknowledgements}
This work is partially funded by Beijing Natural Science Foundation (4252054), Youth Innovation Promotion Association CAS(Grant No.2022132), Beijing Nova Program(20230484276), and CCF-Kuaishou Large Model Explorer Fund (NO. CCF-KuaiShou 2024005).
\clearpage
\setcounter{page}{1}
\maketitlesupplementary
\begin{appendices}
\section{Detailed Proofs}
\paragraph{Relationship between $\beta$/$\gamma$ and $\beta_{new}$/$\gamma_{new}$.}After the magnitude of $\phi(Q_i)$ is scaled by a factor of $a > 1$, we have:
\begin{equation}
\begin{aligned}
    &\gamma_{new}=\frac{a\phi(Q_i)\sum_{m=1}^N\phi(K_m)^T}{N}=a\gamma,\\
    &\beta_{new}-1=\frac{1}{a\phi(Q_i)\sum_{m=1}^N\phi(K_m)^T}=\frac{1}{a}(\beta-1);
\end{aligned}
\end{equation}
Based on the above two equations, we derive the following result:  
\begin{equation}
    \begin{aligned}
        &\gamma_{new}=a\gamma,\\
        &\beta_{new}=\frac{\beta+a-1}{a};
    \end{aligned}
\end{equation}

\paragraph{Proof of $\frac{a\beta}{a+\beta-1}>1$.}Since $\beta>1$ and $a>1$, we have:
\begin{equation}
    (a-1)(\beta-1)>0,
\end{equation}
Expanding the equation, we obtain:  
\begin{equation}
\label{eq:1}
\begin{aligned}
& a\beta-(a+\beta-1)>0, \\
& a\beta>a+\beta-1;
\end{aligned}
\end{equation}
Since $a+\beta>1+1=2>1$, so $a+\beta-1>0$. Based on Eq.~\ref{eq:1}, we have:
\begin{equation}
    \frac{a\beta}{a+\beta-1}>1
\end{equation}

\paragraph{Proof of $p_m>p$.}We define:
\begin{equation}
\begin{aligned}
&A_m=\beta \phi(Q_i)\phi(K_m)^T,\\
&A_n=\beta \phi(Q_i)\phi(K_n)^T;\\
\end{aligned}
\end{equation}
Where we assume that $Q_i$ allocate more attention to $K_m$, thus $A_m>A_n$. Consider a function of \( x \):  
\begin{equation}
    f(x)=\frac{A_m-\gamma x}{A_n-\gamma x};
\end{equation}
The derivative of the function \( f(x) \) can be expressed as:  
\begin{equation}
f'(x) = \frac{\gamma (A_m - A_n)}{(A_n - \gamma x)^2}>0;
\end{equation}

Since we only consider the positive attention scores, we have $\frac{A_n}{\gamma}>\frac{a\beta}{a+\beta-1}>1$, the function \( f(x) \) is monotonically increasing. Thus $p_m=f(\frac{a\beta}{a+\beta-1})>f(1)=p$.

\section{Visualization on Natural Images}
\begin{figure}[t]
    \centering
    \includegraphics[width=0.95\linewidth]{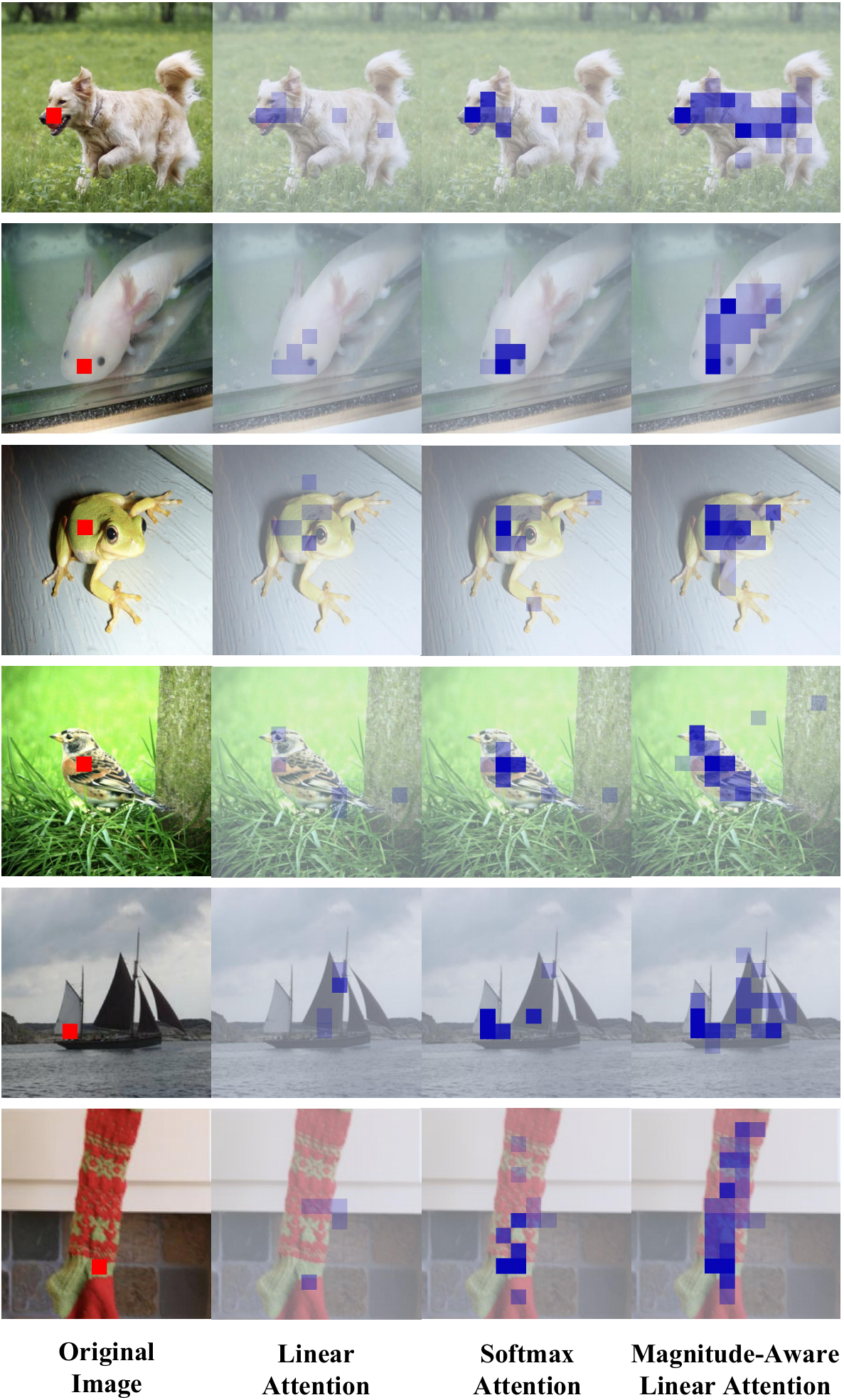}
    \vspace{-2mm}
    \caption{The distribution of attention scores from DeiT-T. Feature corresponding to the red block is used as query.}
    \vspace{-4mm}
    \label{fig:keshi}
\end{figure}

As shown in Fig.~\ref{fig:keshi}, we present the distribution trends of attention scores on natural images. It can be observed that the attention scores of Linear Attention are overly smooth, whereas Softmax Attention is spiky and excessively focuses on local information. Magnitude-Aware Linear Attention effectively balances the characteristics of both.

\end{appendices}

{
    \small
    \bibliographystyle{ieeenat_fullname}
    \bibliography{main}
}

\end{document}